\documentclass[10pt,conference]{IEEEtran}
\usepackage{amsmath,amssymb,amsfonts}
\usepackage{algorithmic}
\usepackage{graphicx}
\usepackage{textcomp}
\usepackage{xcolor}

\usepackage{url}
\usepackage{amsmath}
\usepackage{amsfonts}
\usepackage{comment}
\usepackage{graphicx}
\usepackage{caption}
\usepackage{subcaption}
\usepackage{enumitem}

\usepackage[numbers,sort&compress]{natbib}

\begin{document}

\title{EnsAug: Augmentation-Driven Ensembles for Human Motion Sequence Analysis} 

\author{\IEEEauthorblockN{Bikram De, Habib Irani, Vangelis Metsis}
\IEEEauthorblockA{\textit{Department of Computer Science} \\
\textit{Texas State University}\\
San Marcos, TX, USA \\
\{bikramkumarde, habibirani, vmetsis\}@txstate.edu}}

\maketitle

\begin{abstract}
Data augmentation is a crucial technique for training robust deep learning models for human motion, where annotated datasets are often scarce. However, generic augmentation methods often ignore the underlying geometric and kinematic constraints of the human body, risking the generation of unrealistic motion patterns that can degrade model performance. Furthermore, the conventional approach of training a single generalist model on a dataset expanded with a mixture of all available transformations does not fully exploit the unique learning signals provided by each distinct augmentation type. We challenge this convention by introducing a novel training paradigm, EnsAug, that strategically uses augmentation to foster model diversity within an ensemble. Our method involves training an ensemble of specialists, where each model learns from the original dataset augmented by only a single, distinct geometric transformation. Experiments on sign language and human activity recognition benchmarks demonstrate that our diversified ensemble methodology significantly outperforms the standard practice of training one model on a combined augmented dataset and achieves state-of-the-art accuracy on two sign language and one human activity recognition dataset while offering greater modularity and efficiency. 
Our primary contribution is the empirical validation of this training strategy, establishing an effective baseline for leveraging data augmentation in skeletal motion analysis.
\end{abstract}


\bibliographystyle{ieeetr}

\section{INTRODUCTION}
\begin{figure}
\centering
\includegraphics[width=0.8\linewidth]{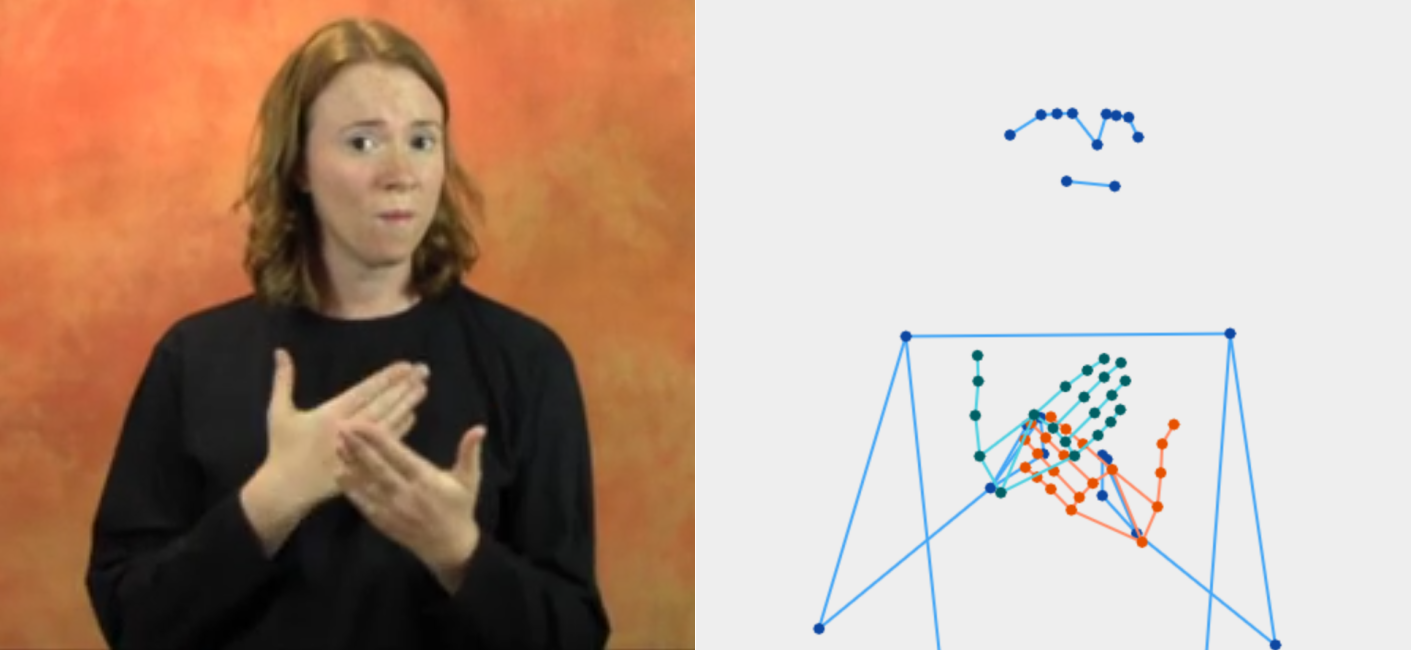}
\caption{\small Example of a person performing a sign in the WLASL dataset \citep{li2020word} and the pose and hands landmarks extracted with MediaPipe \citep{mediapipe}.} \label{fig:sign_lang}
\vspace{-5mm}
\end{figure}

Motion sequence classification, which seeks to interpret human movements from time-series data like skeletal joint trajectories, is a cornerstone of modern human-computer interaction. Key applications, including sign language recognition (SLR) (see Fig.~\ref{fig:sign_lang}) and human activity recognition (HAR), have advanced significantly with the rise of deep learning. In SLR, they enable accessible communication technology for the Deaf and hard-of-hearing communities, real-time translation services, and inclusive education. In HAR, these models play critical roles in healthcare (e.g., fall detection, rehabilitation monitoring), smart surveillance, interactive entertainment, robotics control, and workplace ergonomics \citep{chen2015utd, li2020word}. 

Computational constraints on consumer devices often necessitate a shift from computationally intensive image/video-based models to more efficient landmark-based approaches. By using skeletal keypoints as input, these methods offer robustness against environmental factors such as background clutter and variable lighting, making them well-suited for scalable, real-time systems \citep{li2020word}. However, it is critical to distinguish between \textit{video-based} and \textit{landmark-based} approaches when evaluating performance. Recent foundational models leverage raw RGB frames, large-scale pre-training, or multi-modal text supervision to achieve high accuracy \citep{zhao2024self}. While powerful, these methods incur a prohibitive computational cost and privacy footprint that renders them unsuitable for many edge-computing applications. We target the landmark-based domain, operating on sparse skeletal coordinates that require orders of magnitude less data and compute. We therefore benchmark against other coordinate-based architectures rather than video-pretrained models.

Despite the advantages of landmark-based approaches, large-scale annotated datasets remain scarce, particularly for domain adaptation, gesture variation, and rare-event coverage. While recent egocentric and multimodal datasets \citep{ma2024nymeria} have expanded available resources, domain-specific labeled data for nuanced gestures, regional sign vocabularies, or specialized sensor setups is still limited \citep{li2020word, von2010signum}.

Data augmentation is a widely adopted technique to address this challenge. However, many augmentation strategies for time-series data are direct adaptations from the image domain, such as jittering, scaling, or random noise addition. While these methods increase data quantity, they often fail to respect the inherent geometric structure and kinematic dependencies of the human skeleton. This oversight can generate unrealistic motion artifacts that hinder learning.

To address these challenges, we propose a novel training methodology that strategically combines domain-specific data augmentation with ensemble learning. Our approach involves training an ensemble of deep learning models, where each member is trained on the original dataset augmented by a \textit{single, distinct} geometric transformation. These augmentations (variations in camera depth, spatial shifting, and limb-specific rotations) mimic plausible real-world motion capture variations.

The core hypothesis is that this ensemble of specialists fosters superior model diversity compared to conventional training. By tasking each model with mastering a specific type of variation, we encourage the development of complementary feature representations. The final prediction, aggregated via a simple voting scheme, benefits from this collective expertise, proving more robust and accurate than a single generalist model trained on a mixture of all augmentations. This method also surpasses standard ensembling techniques like bagging, highlighting that the structured diversity introduced by our geometry-aware augmentations is more effective for this domain than the diversity generated by random data sampling.

The main contributions of this paper are as follows:
\begin{itemize}[noitemsep,nolistsep]
    \item We propose and validate a novel training methodology that integrates geometry-aware data augmentation with ensemble learning, demonstrating that training specialized models on distinct augmentations is a highly effective strategy for motion recognition.
    \item We introduce several practical, geometry-aware augmentation techniques for skeletal motion data designed to simulate realistic variations in camera perspective, subject position, and motion dynamics.
    \item Experiments on WLASL, SIGNUM, and UTD-MHAD validate that our ensemble approach significantly outperforms training a single model on a conventionally augmented dataset.
    \item We demonstrate that EnsAug achieves state-of-the-art accuracy among landmark-based approaches, providing an effective training paradigm for motion analysis.
\end{itemize}

We will make data and code available upon acceptance.

\section{RELATED WORK}

\subsection{Data Augmentation for Motion Sequences}
Data augmentation techniques from image classification (cropping, scaling, jittering) \citep{gao2025data} have been adapted for time-series analysis, where methods such as noise injection, magnitude scaling, and time warping improve model robustness \citep{de2024impact}.

However, when applied to skeletal motion data, such generic transformations can be problematic. Human movement is governed by complex biomechanical and kinematic constraints; a simple random perturbation of joint coordinates can easily result in an anatomically impossible pose, injecting unrealistic artifacts into the training process. Recognizing this limitation, research has shifted towards more sophisticated, structure-aware augmentation strategies. A prominent example is \textit{PoseAug} \citep{gong2021poseaug}, a framework that introduces a learnable and differentiable augmentation module. Instead of applying pre-defined transformations, PoseAug is jointly optimized end-to-end with the main pose estimation model. It uses the estimator's feedback to dynamically generate harder and more diverse poses online, while a discriminator module helps ensure biomechanical plausibility. Another influential line of work is \textit{MotionAug} \citep{maeda2022motionaug}, which focuses directly on human motion prediction rather than recognition. MotionAug synthesizes new motion sequences using generative models and inverse kinematics, and then refines them using physics-based imitation learning and debiasing to enforce physical plausibility. Despite strong results, PoseAug and MotionAug introduce significant complexity and computational cost. PoseAug's end-to-end video-based pipeline limits real-time use and is not directly comparable to pure landmark methods. MotionAug's physics-driven synthesis further increases resource requirements, making both approaches less practical for scalable recognition. We propose an alternative: computationally efficient, geometrically-informed offline augmentations combined with an ensemble of models. This modular approach achieves competitive performance without the overhead of complex online generative frameworks.

\subsection{Ensemble Learning for Deep Models}
Ensemble learning rests on the principle that combining multiple models yields more accurate predictions than any single model. Efficacy hinges on member \textit{diversity}: individual models should make uncorrelated errors so that majority voting filters out individual mistakes \citep{breiman1996bagging}.

Traditionally, diversity is induced through methods like bagging, which trains models on different subsets of the data \citep{breiman1996bagging}, or by using varied model architectures. A promising yet less explored strategy for deep learning is to generate diversity by exposing models to systematically different views of the same data. Data augmentation presents a natural mechanism for creating these views. While most studies apply a mixture of augmentations to train a single, robust model, our work investigates an alternative: using each distinct augmentation to train a separate specialist model. This approach treats augmentation not just as a means of data expansion, but as a deliberate strategy for cultivating model diversity within an ensemble framework \citep{ganaie2022ensemble}.

\subsection{Landmark-based Motion Recognition}
Sign language recognition (SLR) and human activity recognition (HAR) historically relied on processing raw video frames with CNNs. While effective, image-based systems are computationally demanding and sensitive to background clutter and occlusion \citep{li2020word}. The field has consequently shifted toward landmark-based approaches, where pose estimation pipelines such as MediaPipe or OpenPose extract compact skeletal keypoint sequences \citep{mediapipe}. This low-dimensional representation is efficient and inherently robust to visual noise, enabling sequence models like LSTMs and Transformers for real-time recognition. Both SLR \citep{von2010signum} and HAR \citep{chen2015utd} using landmarks have shown strong results, yet limited annotated data in both domains motivates advanced training paradigms such as ours.

\section{METHODOLOGY}
\begin{figure*}
    \centering
    \includegraphics[width=0.9\linewidth]{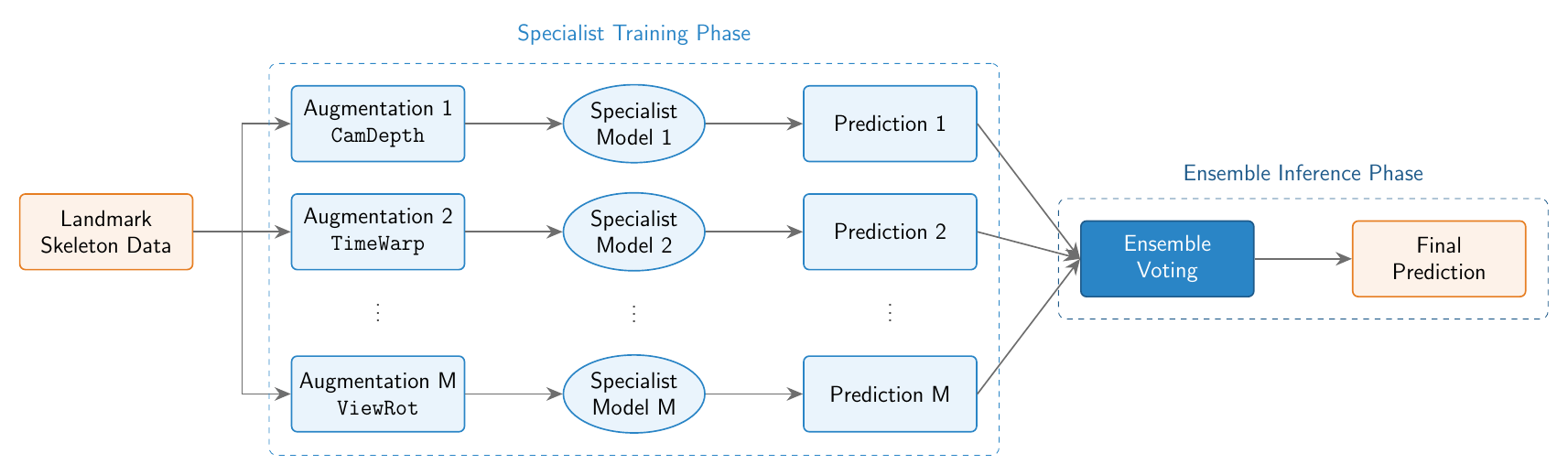}
    \vspace{-2mm}
    \caption{\small EnsAug framework pipeline.}
    \vspace{-3mm}
    \label{fig:framework}
\end{figure*}

EnsAug operates on 3D skeletal landmark sequences sourced from motion capture (e.g., UTD-MHAD via Kinect) or extracted from video using pose estimation toolkits such as MediaPipe. The method uses data augmentation to create an ensemble of specialized models (Figure~\ref{fig:framework}).

The process involves two main phases. First, during the \textit{specialist training} phase, we generate $M$ distinct copies of the original training dataset. Each copy is transformed using a single, unique geometry-aware augmentation technique. We then train $M$ separate deep learning models, where each model $\mathcal{M}_i$ is trained exclusively on the dataset treated with the $i$-th augmentation. This yields an ensemble of $M$ specialist models. Second, during the \textit{ensemble prediction} phase, a test sample is fed to all $M$ trained specialists, and their individual predictions are aggregated using a voting strategy to produce a final, robust classification.

\subsection{Geometry-Aware Augmentation}
We define a motion sequence as a set of landmarks $\mathbf{L}$, where $\mathbf{L}_{t,j} = (x_{t,j}, y_{t,j}, z_{t,j})$ represents the 3D coordinates of joint $j$ at timestep $t$. Each augmentation transforms $\mathbf{L}$ into $\mathbf{L}'$ by simulating realistic variations.

\subsubsection{Camera Depth Variation (CamDepth)}
This technique simulates the subject being closer to or farther from the camera. We apply a uniform scaling factor $s$ to the z-coordinate of all landmarks across all frames, preserving the relative internal geometry of the skeleton.
\vspace{-2mm}
\begin{align}
\forall\, t, j: \quad \mathbf{L}'_{t,j} = (x_{t,j}, y_{t,j}, s \cdot z_{t,j})
\end{align}
where $s \in \mathbb{R}^+$ is a randomly sampled scaling factor.

\subsubsection{Temporal Depth Change (TempDepth)}
To model scenarios where a subject moves towards or away from the camera during the action, we apply a time-varying scaling factor $s_t$ to the z-coordinate.
\vspace{-2mm}
\begin{align}
\forall\, t, j: \quad \mathbf{L}'_{t,j} = (x_{t,j}, y_{t,j}, s_t \cdot z_{t,j})
\end{align}
The sequence of scaling factors $\{s_t\}_{t=1}^T$ is a smooth function, such as a linear ramp or a single cycle of a sine wave, simulating continuous movement.

\subsubsection{Horizontal \& Vert. Shifting (HV-Shift)}
This augmentation models lateral or vertical shifts of the subject within the camera's frame. We add a displacement $(\Delta x_t, \Delta y_t)$ to the x and y coordinates.
\vspace{-2mm}
\begin{align}
\forall\, t, j: \quad \mathbf{L}'_{t,j} = (x_{t,j} + \Delta x_t, y_{t,j} + \Delta y_t, z_{t,j})
\end{align}
The shifts can be constant for all frames or can vary smoothly over time to simulate drifting movements.

\subsubsection{Hand Size Variation (HandSize)}
To account for natural differences in anthropometry, we simulate variations in hand size by scaling the hand landmarks relative to a center point, typically the wrist.
\vspace{-2mm}
\begin{align}
\forall\, t, j \in \{\text{hand joints}\}: \quad \mathbf{L}'_{t,j} = \mathbf{w}_t + \alpha (\mathbf{L}_{t,j} - \mathbf{w}_t)
\end{align}
where $\mathbf{w}_t$ is the wrist landmark at time $t$, and $\alpha$ is the scaling factor.

\subsubsection{Viewpoint Rotation (ViewRot)}
This simulates capturing the motion from a different camera angle by rotating the entire skeleton around a central pivot.
\vspace{-2mm}
\begin{align}
\forall\, t, j: \quad \mathbf{L}'_{t,j} = \mathbf{R} (\mathbf{L}_{t,j} - \mathbf{c}_t) + \mathbf{c}_t
\end{align}
where $\mathbf{R} \in SO(3)$ is a rotation matrix for a randomly sampled axis and angle, and $\mathbf{c}_t$ is the skeleton's centroid at time $t$.

\subsubsection{Finger Articulation (FingerFold)}
This simulates the natural curling or folding of fingers by applying rotations at the key finger joints: the Metacarpophalangeal (MCP, knuckle), Proximal Interphalangeal (PIP, middle), and Distal Interphalangeal (DIP, tip) joints.
\vspace{-2mm}
\begin{align}
\forall\, t: \quad \mathbf{f}'_{t,1} &= \mathbf{f}_{t,0} + \mathbf{R}_{\mathrm{MCP}}(\mathbf{f}_{t,1} - \mathbf{f}_{t,0}) \\
\mathbf{f}'_{t,2} &= \mathbf{f}'_{t,1} + \mathbf{R}_{\mathrm{PIP}}(\mathbf{f}_{t,2} - \mathbf{f}_{t,1}) \\
\mathbf{f}'_{t,3} &= \mathbf{f}'_{t,2} + \mathbf{R}_{\mathrm{DIP}}(\mathbf{f}_{t,3} - \mathbf{f}_{t,2})
\end{align}
where $\mathbf{f}_{t,k}$ are the landmarks along a finger chain at time $t$, and each $\mathbf{R}$ is a rotation matrix parameterized by a folding angle, ensuring biomechanical plausibility.

\subsubsection{Elbow-driven Hand Displ. (ElbowDisp)}
To emulate forearm flexion and extension, we apply a displacement vector $\mathbf{d}_t$ to all hand landmarks. This shifts the entire hand assembly inward or outward relative to the torso.
\vspace{-2mm}
\begin{align}
\forall\, t, j \in \{\text{hand joints}\}:  \quad \mathbf{L}'_{t,j} = \mathbf{L}_{t,j} + \mathbf{d}_t
\end{align}
where $\mathbf{d}_t \in \mathbb{R}^3$ is a time-dependent displacement vector.

\subsubsection{Time Warping (TimeWarp)}
This technique alters the temporal aspect of a sequence to simulate variations in performance speed. A warping function $\tau(t')$ maps the new timeline to the original, effectively stretching or compressing parts of the motion.
\vspace{-2mm}
\begin{align}
\forall\, j: \quad \mathbf{L}'_{t',j} = \mathbf{L}_{\tau(t'),j}
\end{align}
Linear interpolation is used for non-integer values of $\tau(t')$ to ensure a smooth sequence.

\subsection{Ensemble Aggregation}
Once the $M$ specialist models $\{\mathcal{M}_1, \ldots, \mathcal{M}_M\}$ are trained, their individual predictions for a given test instance $x$ must be aggregated to form a single, final decision. We employ majority voting due to its simplicity, efficiency, and robust performance in our experiments.

In this approach, often called Hard Voting, each of the $M$ classifiers casts one vote for its predicted class. The final prediction of the ensemble, $H(x)$, is simply the class that receives the highest number of votes. This is formally expressed as:
\vspace{-3mm}
\begin{equation}
H(x) = \arg\max_{c} \sum_{i=1}^M \mathbb{I}\{ h_i(x) = c \}
\label{eq:hard_voting}
\end{equation}
where $h_i(x)$ is the prediction of model $\mathcal{M}_i$ for the input $x$, and $\mathbb{I}\{\cdot\}$ is the indicator function.

An alternative is Probability Averaging (Soft Voting), which averages output class probabilities. We found majority voting consistently effective; it serves as the primary aggregation method in this work.

Standard training combines all data augmentations into a single generalist model. We hypothesize that distinct geometric transformations can induce contradictory gradient updates in a shared weight space. For example, learning invariance to Scale (which alters global size) may interfere with learning invariance to Viewpoint Rotation (which alters relative joint angles). By isolating these transformations into specialist models, EnsAug allows each network to learn distinct, non-conflicting invariant features, which are then effectively recombined via the ensemble vote.

\section{EXPERIMENTS}
We evaluate EnsAug on three datasets: WLASL and SIGNUM for sign language recognition, and UTD-MHAD for human action recognition. Experiments address two questions: 1) How does each geometry-aware augmentation compare to a non-augmented baseline? 2) Does the ensemble of specialists outperform the baseline and the ``Generalist'' model?

For UTD-MHAD (full-body actions), we adapted augmentations accordingly: `HandSize' became `PersonSize' (scaling the entire skeleton), `HV-Shift' became `ShiftBody', and hand-specific augmentations (`FingerFold', `ElbowDisp') were omitted.

\subsection{Datasets}

\subsubsection{WLASL}
The Word-Level American Sign Language (WLASL) dataset \citep{li2020word} is a large-scale collection of isolated ASL signs derived from online videos. Following common practice, we extract 3D landmarks for both hands (42 landmarks in total, 126 features) using the MediaPipe Holistic pipeline \citep{mediapipe}. All sequences are normalized to 80 timesteps (trimming or zero-padding) and centered across subjects. We adhere to the official dataset splits, which partition the data by gloss into training, validation, and test sets. We report results on the two most common subsets:
\begin{itemize}[noitemsep,nolistsep]
    \item \textbf{WLASL-100:} Contains the 100 most frequent signs, comprising 2,038 videos from 97 unique signers.
    \item \textbf{WLASL-300:} A larger subset with 300 signs, containing 5,117 videos from 109 signers.
\end{itemize}

\subsubsection{SIGNUM}
The SIGNUM dataset \citep{von2010signum} consists of 450 basic signs from German Sign Language (DGS) performed by 25 different signers. We process this dataset identically to WLASL, extracting 3D hand landmarks via MediaPipe and normalizing all sequences to a length of 80 timesteps. For a robust signer-independent evaluation, we partition the dataset by signer ID as follows:
\begin{itemize}[noitemsep,nolistsep]
    \item \textbf{Training set:} 14 signers (IDs 1-14) -- 6,300 sequences.
    \item \textbf{Validation set:} 4 signers (IDs 15-18) -- 1,800 sequences.
    \item \textbf{Test set:} 7 signers (IDs 19-25) -- 1,901 sequences.
\end{itemize}
\subsubsection{UTD-MHAD}
The UT-Dallas Multimodal Human Action Dataset (UTD-MHAD) \citep{chen2015utd} contains 27 distinct actions performed by 8 subjects. For our experiments, we use the 3D skeletal data captured by a Kinect sensor, which provides the 3D coordinates of 20 joints. The dataset contains 861 valid sequences after removing corrupted samples. To ensure a subject-independent evaluation, we split the data by subject ID:
\begin{itemize}[noitemsep,nolistsep]
    \item \textbf{Training set:} 5 subjects (IDs 1-5) -- 539 sequences.
    \item \textbf{Test set:} 3 subjects (IDs 6-8) -- 322 sequences.
\end{itemize}
Given the small number of subjects, we follow a direct training-to-test evaluation protocol and do not use a separate validation set for this dataset.

\subsection{Motivation: Limitations of Generic Augmentations}
We first evaluated generic time-series augmentations \citep{de2024impact} (jittering, uniform scaling) that do not account for skeletal geometry. Table~\ref{tab:traditional_aug} shows that these augmentations yield only modest gains; even an ensemble (`Ensemble' row) offers limited improvement. Generic augmentations that treat all channels uniformly can disrupt kinematic relationships, motivating our geometry-aware approach.

\begin{table}
\caption{\small Performance (Accuracy \%) of the baseline model trained with traditional, generic time-series augmentation techniques. The `Ensemble' row shows the result of combining these models with majority voting.}\label{tab:traditional_aug}
\centering
\begin{tabular}{|l|c|c|c|c|}
\hline
Model & Wlasl100 & Wlasl300 & Signum & MHAD \\ \hline
Jitter     & .610 & .452 & .846 & .559 \\
Scale      & .619 & .468 & .838 & .562 \\
MagWarp    & .526 & .423 & .830 & .509 \\
TimeWarp   & .663 & .537 & .852 & .512 \\
WinWarp    & .635 & .543 & .900 & .584 \\
WinScale   & .639 & .539 & .889 & .612 \\
Ensemble   & .663 & .569 & .904 & .615 \\
\hline
\end{tabular}
\vspace{-3mm}
\end{table}

\subsection{Implementation Details}
All experiments were conducted using the PyTorch framework on Azure Standard\_NC4as\_T4\_v3 instances, each equipped with an NVIDIA Tesla T4 GPU.

\begin{figure}
\centering
\includegraphics[width=.43\columnwidth, angle=90]{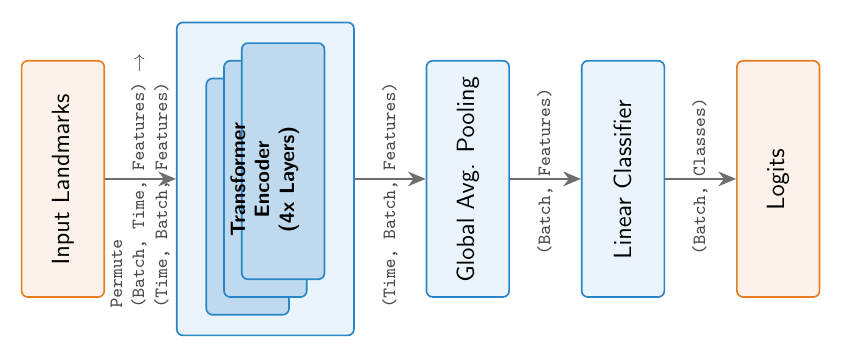}
\caption{\small The Transformer architecture used for each specialist model. The input landmark sequence is passed through a stack of encoder layers, followed by global average pooling over the time dimension to produce a fixed-size representation for the final classifier.}
\vspace{-3mm}
\label{fig:model_arch}
\end{figure}

\noindent\textbf{Model Architecture.} Our base model for all experiments is a standard Transformer encoder architecture, as implemented in PyTorch and illustrated in Figure~\ref{fig:model_arch}. The model consists of 4 encoder layers, with 9 attention heads each, a feedforward hidden dimension of 256, and a dropout rate of 0.1. The input to the model is the sequence of landmark coordinates, and the final representation for classification is taken from the output of the last timestep of the encoder.

\noindent\textbf{Training Protocol.} We trained each model using the Adam optimizer with a learning rate of $1 \times 10^{-4}$ and a batch size of 16. The loss was calculated using the standard Cross-Entropy Loss function. For each training run, we trained for a maximum of 400 epochs and used an early stopping strategy based on a dedicated validation set. The model weights that achieved the highest validation accuracy during the training process were saved and used for the final evaluation on the test set.

\begin{figure}[t]
\centering
\includegraphics[width=.8\linewidth]{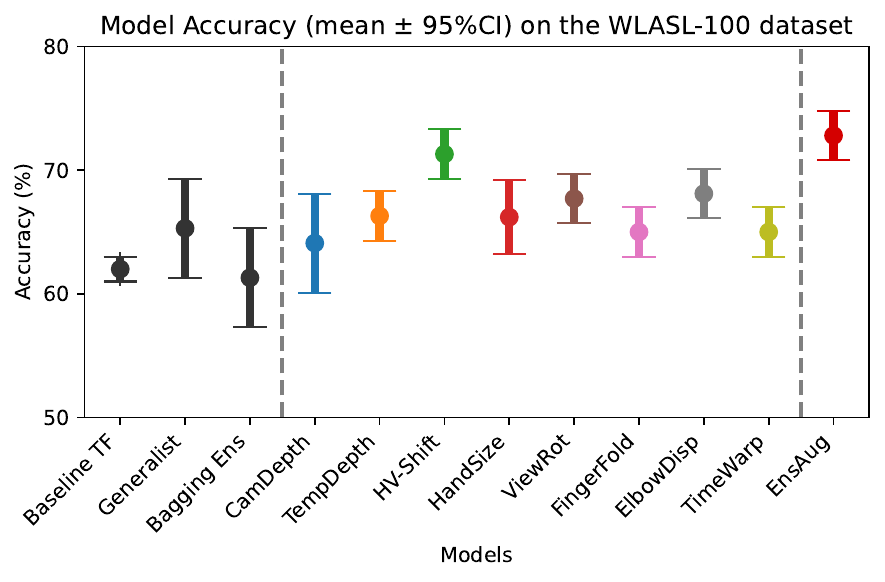}
\vspace{-3mm}
\caption{\small Model results on the WLASL-100 dataset. All metrics are percentages (mean $\pm$ 95\% CI).} \label{fig:WLASL-100}
\vspace{-3mm}
\end{figure}

\begin{figure}[t]
\centering
\includegraphics[width=.8\linewidth]{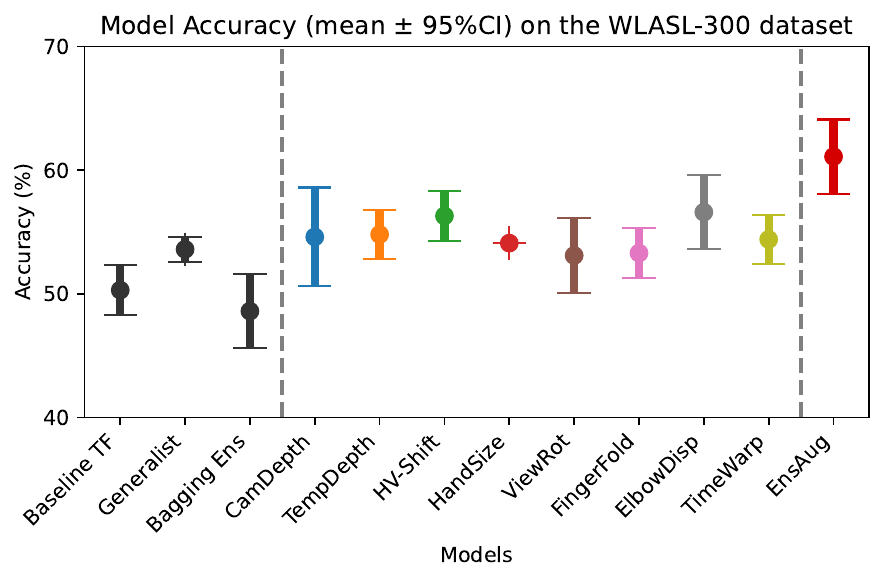}
\vspace{-3mm}
\caption{\small Model results on the WLASL-300 dataset. All metrics are percentages (mean $\pm$ 95\% CI).} \label{fig:WLASL-300}
\end{figure}

\begin{table}[b]
\vspace{-3mm}
\caption{\small Augmentation patterns and ranges.}
\label{tab:augpatterns}
\centering
\resizebox{\columnwidth}{!}{
\begin{tabular}{|l|l|l|}
\hline
\textbf{Model} & \textbf{Pattern} & \textbf{Hyperparameter Range} \\
\hline
CamDepth & Near                      & Distance: 0.7            \\
         & Far                       & Distance: 1.3            \\
\hline
TempDepth & Far to near              & Distance: (0.5, 1.3)     \\
          & Near to far              & Distance: (0.7, 1.5)     \\
          & Near far near            & Distance: (0.6, 1.4)     \\
\hline
HV-Shift  & Linear shift             & Distance: $(-0.15, 0.15)$\\
          &                          & amplitude\_x: 0.10, \\ 
          &                          & amplitude\_y: 0.08, \\
          & Sine shift                        & frequency: 1.5 \\
\hline
HandSize  & Large Hands              & Scale factor: 1.3        \\
          & Small Hands              & Scale factor: 0.8        \\
\hline
ViewRot   & Yaw rotation along center& Degree: $(-45, +45)$     \\
\hline
FingerFold & Gradual finger fold    & folding\_progression: \\
           & per timestep           & [0.2, 0.4, 0.6, 0.8] \\
\hline
ElbowDisp  & Inward                  & Intensity: 0.4           \\
           & Outward                 & Intensity: 0.3           \\
\hline
TimeWarp   & Mild to moderate time warp      & $\sigma = 0.1$ and $0.05$, knot = 4 \\
\hline
\end{tabular}
}
\vspace{-5mm}
\end{table}

\noindent\textbf{Augmentation Parameters.} The hyperparameters for each augmentation were chosen to generate plausible variations. For example, for the \textbf{HV-Shift} augmentation, the shift distance was typically sampled from a range of $[-0.15, 0.15]$ relative to the normalized coordinate space. Table~\ref{tab:augpatterns} provides the full list of augmentation patterns and ranges used.

\subsection{Results}

\begin{table}[t]
\centering
\caption{\small Comparison with model performance (in \% accuracy) from previous works.
Pose-GRU and Pose-TGCN can be found in \citep{li2020word},
GCN-BERT can be found in \citep{tunga2021pose},
Procrustes-DTW (P-DTW) can be found in \citep{de2025geometric},
Spoter can be found in \citep{bohavcek2022sign},
EnsAug (Ours) represents our ensemble augmentation framework.}\label{tab:sota}
\begin{subtable}{.5\columnwidth}
    \centering
    \caption{WLASL datasets.}
    \label{tab:wlasl_results}
    \resizebox{\linewidth}{!}{%
    \begin{tabular}{|l|c|c|}
    \hline
    \textbf{Model}      & W-100 & W-300 \\ \hline
    Pose-GRU           & 46.51    & 33.68    \\
    Pose-TGCN          & 55.43    & 38.32    \\
    GCN-BERT           & 60.15    & 42.18    \\
    P-DTW     & 61.10    & 51.85    \\
    Spoter             & 46.51    & 33.68    \\
    \textbf{EnsAug} & \textbf{72.80} & \textbf{61.10} \\ \hline
    \end{tabular}%
    }
\end{subtable}%
\hfill 
\begin{subtable}{.46\columnwidth}
    \centering
    \caption{\centering SIGNUM and MHAD.}
    \label{tab:signum_mhad_results}
    \resizebox{\linewidth}{!}{%
    \begin{tabular}{|l|c|c|}
    \hline
    \textbf{Model}      & SIGNUM         & MHAD           \\ \hline
    P-DTW    & 90.20          & 64.90          \\
    \textbf{EnsAug} & \textbf{92.70} & \textbf{67.60} \\ \hline
    \end{tabular}%
    }
\end{subtable}

\end{table}

\begin{figure}[t]
\centering
\includegraphics[width=.8\linewidth]{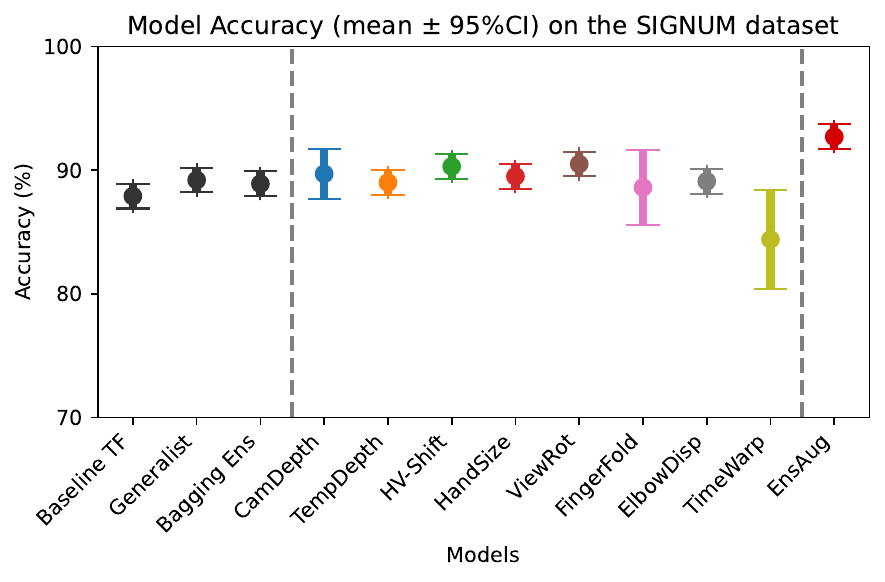}
\vspace{-3mm}
\caption{\small Model results on the SIGNUM dataset. All metrics are percentages (mean $\pm$ 95\% CI).} \label{fig:SIGNUM}
\vspace{-3mm}
\end{figure}

\begin{figure}
\centering
\includegraphics[width=.8\linewidth]{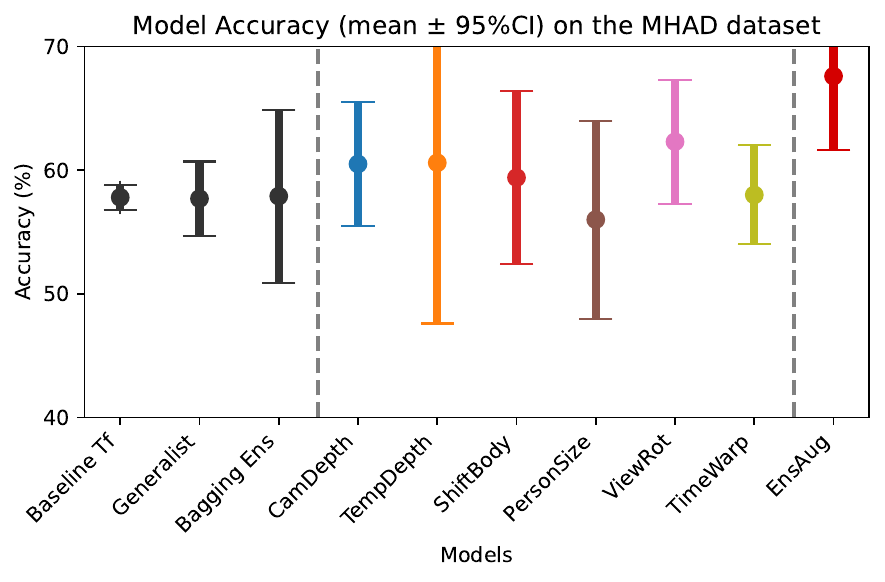}
\vspace{-3mm}
\caption{\small Model results on the MHAD dataset. All metrics are percentages (mean $\pm$ 95\% CI).} \label{fig:MHAD}
\vspace{-3mm}
\end{figure}

Figures~\ref{fig:WLASL-100}--\ref{fig:MHAD} present results on all datasets. All reported metrics are the mean over five runs, and the uncertainty is expressed as the 95\% confidence interval. In each figure, ``Baseline TF'' refers to the Transformer model trained on the original, non-augmented data. The ``Generalist'' refers to the model with random augmentations applied in each batch. The ``Bagging Ens'' refers to a model where we created a bootstrapped dataset with replacement each time instead of an augmented dataset and then created an ensemble of the individual results. The subsequent labels show the performance of individual specialist models, each trained with a single augmentation. The final row, \textbf{EnsAug (Ours)}, shows the performance of our complete augmentation-driven ensemble framework using majority voting.

Across all datasets and metrics, our proposed \textit{EnsAug} framework consistently and significantly outperforms both the baseline model and every individual specialist model. This result strongly supports our central hypothesis that an ensemble of specialists, each focused on a different data variation, yields a more robust and accurate system than any single model alone. Among the individual augmentations, `HV-Shift' (and its full-body counterpart `ShiftBody') consistently proves to be one of the most effective single transformations.



\subsection{Ablation Studies}

\noindent\textbf{Error Diversity.} To quantify specialist complementarity, we computed the pairwise Jaccard Index of misclassified samples between all model pairs. Lower overlap indicates more diverse error patterns. Table~\ref{tab:jaccard} shows that SIGNUM exhibits the lowest overlap (mean 0.37), indicating higher specialist diversity, while WLASL-300 shows the highest (mean 0.63). This confirms that specialists trained on different geometric projections learn to correct distinct subsets of errors.

\begin{table}
\centering
\caption{\small Pairwise error overlap (Jaccard Index) between specialists. Lower values indicate more diverse error patterns.}
\label{tab:jaccard}
\vspace{-2mm}
\begin{tabular}{|l|c|c|c|}
\hline
\textbf{Dataset} & \textbf{Min} & \textbf{Mean} & \textbf{Max} \\ \hline
SIGNUM     & 0.28 & 0.37 & 0.43 \\
MHAD       & 0.44 & 0.53 & 0.59 \\
WLASL-100  & 0.50 & 0.58 & 0.68 \\
WLASL-300  & 0.59 & 0.63 & 0.66 \\ \hline
\end{tabular}
\vspace{-3mm}
\end{table}

\begin{figure}
\centering
\includegraphics[width=.7\columnwidth]{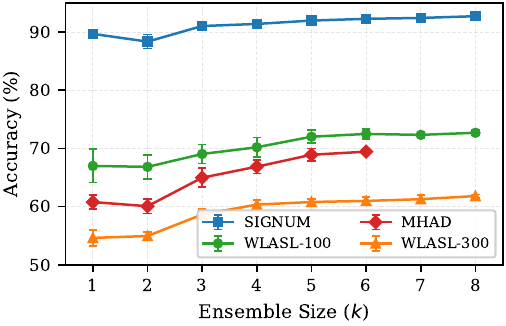}
\vspace{-2mm}
\caption{\small Accuracy vs.\ ensemble size $k$ (mean $\pm$ 95\% CI). MHAD uses $M{=}6$ augmentations; others use $M{=}8$.}
\label{fig:enssize}
\vspace{-3mm}
\end{figure}

\noindent\textbf{Ensemble Size.} We evaluated mean accuracy over all $\binom{M}{k}$ specialist subsets for $k=1\ldots M$. Figure~\ref{fig:enssize} shows accuracy increases with $k$, with diminishing returns beyond $k \approx 5$. For instance, SIGNUM improves from 89.7\% ($k{=}1$) to 92.7\% ($k{=}8$), but already reaches 92.0\% at $k{=}5$. This supports using all specialists while confirming that even smaller ensembles provide substantial gains.

\section{DISCUSSION}
The results validate our core hypothesis: strategically cultivating model diversity through specialized training outperforms training a single generalist model.

The different baselines underscore the importance of our approach. The ``Generalist'' model with random augmentations confirms that augmentation alone is insufficient for optimal accuracy, while ensembling without augmentation is similarly suboptimal. \textit{EnsAug} combines both strategies, achieving superior accuracy and state-of-the-art results in landmark-based approaches. The comparison with the ``Generalist'' baseline reveals key insights about learning from geometric data. While standard practice aggregates all augmentations into a single training loop, our results suggest the existence of a \textit{Geometric Augmentation Conflict}. When a single model is tasked with learning invariance to multiple, distinct geometric transformations simultaneously (e.g., global Scale invariance versus localized Viewpoint Rotation), the gradient updates may become antagonistic, leading to interference in the shared weight space. By decoupling these transformations into separate specialists, EnsAug allows each network to maximize its feature extraction capability for a specific geometric view without interference. Thus, the innovation of \textit{EnsAug} lies not in the voting mechanism itself, but in the architectural decision to use ensembling as a solution to geometric conflict, proving that the isolation of geometric constraints is as critical as the augmentation itself.

This performance gain addresses the critique that ensembles rely solely on parameter aggregation. A standard ensemble approach, such as Bagging (bootstrap aggregating), relies on random data subsampling to induce diversity. As shown in our results, EnsAug significantly outperforms the Bagging baseline. This distinction is crucial: the novelty of EnsAug lies not in the voting mechanism, but in the use of distinct geometric projections to force feature diversity. As shown in Table~\ref{tab:jaccard}, the pairwise error overlap between the different specialists is relatively low, confirming that specialists trained on different geometries learn to classify different subsets of hard samples correctly. Our contribution establishes a framework where ensemble members are engineered as geometric specialists for skeletal motion analysis.

An analysis of the individual specialist models reveals further insights. On the sign language datasets, `HV-Shift' frequently emerges as the top-performing single augmentation. This aligns with the real-world observation that signers often have minor variations in their lateral signing space. Similarly, on the UTD-MHAD dataset, `ViewRot' proves highly effective, mirroring how subjects naturally reorient themselves during complex full-body activities. However, no single augmentation is universally optimal. The ensemble succeeds by harnessing the collective knowledge of all specialists rather than relying on a single ``best'' augmentation. That the ensemble outperforms even the strongest specialist (e.g., `HV-Shift') confirms that models learn complementary features, allowing majority voting to correct individual errors.

A key advantage is parallelization: since specialists train independently, wall-clock time equals that of a single model given sufficient GPUs. Our base model is a lightweight Transformer operating on sparse vector inputs (126 coordinates/frame). Consequently, the aggregate FLOPs of an 8-member EnsAug ensemble remain significantly lower than a single pass of a standard video-based backbone (e.g., ResNet-3D).

Finally, to position our work within the broader field, we compared `EnsAug' to other landmark-only state-of-the-art methods. As shown in Table \ref{tab:sota}, our framework achieves a new state-of-the-art on all evaluated datasets, although for the Signum and UTD-MHAD datasets, the published landmark-only results are limited to only one prior publication. Most prior models are limited to sign language recognition only and, unlike our model, do not extend to broader motion sequence analysis. Combined with EnsAug's simplicity (a standard Transformer and streamlined training), this demonstrates that optimized training strategies can rival more complex architectures. While methods like PoseAug and MotionAug show promise, their high complexity and computational cost, reliance on video, or domain-specific pipelines make them impractical for efficient real-time motion recognition.


\section{CONCLUSION}
We introduced and validated a training paradigm called Augmentation-Driven Ensembles. Training an ensemble of specialist models, where each model learns from a single, distinct geometry-aware augmentation, is an effective strategy for human motion analysis. The resulting framework, \textit{EnsAug}, achieves state-of-the-art performance on multiple benchmark datasets using a simple, modular, and computationally efficient parallel training strategy. This work establishes a new baseline for leveraging data augmentation in motion recognition.


\bibliography{references}

\end{document}